\documentclass[]{style/ceurart}
\usepackage{listings}
\usepackage{hyperref}
\usepackage[table]{xcolor}
\usepackage{booktabs}
\usepackage{tikz}
\usetikzlibrary{shapes.geometric, arrows, positioning}
\begin{document}

\copyrightyear{2026}
\copyrightclause{Copyright for this paper by its authors.
  Use permitted under Creative Commons License Attribution 4.0
  International (CC BY 4.0).}

\conference{CLEF 2026: Conference and Labs of the Evaluation Forum, September 21-24, 2026, Jena, Germany}

\title{DS@GT ARC at LongEval: Citation Integrity and Factual Grounding in Scientific QA}

\author[1]{Brandon Michaels}[
    orcid=0009-0008-2056-6884,
    email=bmichaels6@gatech.edu,
]
\cormark[1]

\author[1]{Brendon Johnson}[
    orcid=0009-0002-1777-2456,
    email=bjohnson436@gatech.edu,
]
\cormark[1]

\address[1]{Georgia Institute of Technology, North Ave NW, Atlanta, GA 30332}
\cortext[1]{Corresponding author.}

\begin{abstract}
This paper describes DS@GT ARC's submission to the CLEF 2026 LongEval Task 4 on Retrieval-Augmented Generation (RAG). In this submission, we examine a divergence between traditional natural language evaluation metrics and citation integrity as applied to RAG QA systems. We evaluate a corrective pipeline using Corrective RAG (CRAG) and CiteFix against baseline and frontier model benchmark RAG QA scores. While frontier models maximized answer relevance and fluency scores, our RAGAs LLM-as-judge diagnostics indicate that frontier models would correctly identify relevant documents without using their context in answer generation. Conversely, by filtering chunks pre-generation and enforcing strict entailment of generated claims to the cited material post-generation, our corrective pipeline marginally improved citation faithfulness and answer grounding. We propose that evaluation of trustworthy RAG QA requires metrics that reward strict answer grounding.
\end{abstract}

\begin{keywords}
  Retrieval-Augmented Generation \sep
  Scientific Question Answering \sep
  Large Language Models \sep
  Attributed Text Generation
\end{keywords}

\maketitle


\section{Introduction}

Scientific knowledge evolves through both continual research and publication of new papers that update previous empirical results, propose novel approaches, or replace old knowledge altogether. While this has led to the discovery of many new research findings, it also has consequential implications for RAG-related question answering tasks. With rapid developments in LLM technology, information can become quickly outdated or even become a complete distraction during question answering or information retrieval tasks \cite{lewis2020rag}. While RAG is an intuitive choice for answering scientific questions since it can base its output on the content of source documents instead of being limited to model memory, it still has ample retrieval related mistakes prompting for stronger, grounded QA. Such mistakes include retrieval of irrelevant documents or paragraphs, missing evidence, wrong document citations, and retrieval of inaccurate or outdated information. These issues become particularly relevant in the domain of science because an answer must be factually grounded, empirically backed by research documents, accurate and currently relevant, and the evidence used must justify the claims.

The 2026 LongEval-RAG \cite{lncs_overview_longeval_2026, ceur_overview_longeval_2026} task considers this issue in relation to temporally evolving scientific documents from the Connecting Repositories (CORE) scientific literature corpus \cite{Knoth2023-zi}. For every query, the task organizers present a set of ten documents to consider, and require the participants to output an answer in natural language form, supported by references to the documents used to generate an evidence based answer. Although this inherently simplifies the initial retrieval process by narrowing the candidate set to a smaller region of documents before answer generation, the task organizers included irrelevant documents which still required evidence related filtering. Even within a chunked document, not all information can be considered relevant or highly relevant, which still makes this RAG task meaningful in both natural language answer quality and faithfulness.

As the task supplied a set of pre-retrieved documents for each query, we focused our architecture on post-retrieval optimization of grounded answer generation. To address the challenge of distractor documents, we compare a hybrid chunk retrieval baseline against a strict corrective pipeline. The corrective architecture utilizes Corrective RAG (CRAG) \cite{yan2024crag} to filter chunks prior to generation, and CiteFix \cite{maheshwari2025citefix} to enforce strict claim attribution by pruning ungrounded citations post-generation.

Apart from the automated pipelines, we also examine two manually curated runs that use state-of-the-art frontier models (GPT-5.5 Thinking and Claude Opus 4.7 Thinking). While these approaches achieved high scores on standard NLP metrics (ROUGE, BERT) and LLM judged answer relevancy, our RAGAs-based faithfulness evaluation reveals that these same frontier models often fail to properly ground answers in cited materials. The remainder of this paper outlines the related work, details our methodologies, and highlights the gap between generative fluency and citation integrity in our evaluation results.

\section{Related Work}
Standard RAG frameworks treat source documents as static repositories of knowledge. However, document corpora often exhibit longitudinal context drift, where knowledge mutates over time \cite{keller2024eval}. This poses a challenge to long-context extraction, where temporal misalignment can contaminate retrieval. Traditional RAG operates under a naive assumption that retrieved contexts are inherently trustworthy, leaving generators vulnerable to retrieval failures. Corrective RAG (CRAG) introduced an active, intermediate evaluation layer that attempts to sanitize the retrieved context before generation \cite{yan2024crag}. Another challenge for RAG systems is the strict attribution of sources. At generation, an LLM can generate coherent text using supplied context but include ungrounded statements or inaccurate citations \cite{liu2023evaluating}. Recent efforts to address this failure mode in trustworthy RAG focus on post-hoc citation mapping to force explicit citation of generated claims \cite{gao-etal-2023-enabling, maheshwari2025citefix}.

\section{Task and Experimental Methodology}

\subsection{CORE Scientific Document Corpus}
The LongEval 2026 Task 4 (LongEval-RAG) dataset provides a textual query alongside 10 pre-retrieved document IDs from the CORE corpus. These 10 retrieved documents are meant to represent the knowledge boundary for generating the answer. This constraint tests a RAG system's ability to filter evidence, isolate distractors, and successfully produce extractive answers. For each document ID, paper title, abstract, and full text fields are made available.

Generative pipelines produced for the LongEval-RAG task must produce outputs aligned with a strict JSONL schema. For each query, systems ingest a JSON containing the query and list of pre-retrieved document IDs. The pipeline processes this input and appends an answer including the generated response and cited documents in the pre-retrieved array. An example of the format is provided in Figure \ref{fig:json-schema}.
\begin{figure}[h]
\centering
\begin{lstlisting}[basicstyle=\ttfamily\small, frame=single, numbers=left, numberstyle=\tiny, xleftmargin=2em]
{
  "metadata": {
    "team_id": "LongEval DS@GT",
    "run_id": "example-run",
    "narrative_id": "aa42e210a...",
    "narrative": "How can a device avoid futile access..."
  },
  "references": [275699672, 275699915, 122371639, ...],
  "answer": [
    {
      "text": "A device can avoid futile NR access by using...",
      "citations": [0, 1] 
    }
  ]
}
\end{lstlisting}
\caption{Expected JSONL schema for LongEval-RAG submissions. Note that the \texttt{citations} array relies on local positional indices mapping back to the \texttt{references} array, rather than the global document IDs in the generated text.}
\label{fig:json-schema}
\end{figure}
\subsection{Candidate Processing and Retrieval}
Given that an initial set of candidate documents is provided by the organizers, we do not aim to create a global corpus retriever. Instead, our architecture focuses on post-retrieval optimization to rank and sanitize the provided payload.

We first segment documents into overlapping semantic chunks. Each chunk is enriched with its parent document's metadata to enable proper attribution. We then execute a second-stage retrieval over these chunks using a hybrid retriever. Our retriever uses a combination of BM25 and BGE (bge-base-en-v1.5) dense vector embeddings to rank the chunks. The top-$k$ chunks based on the input query are fed into the context window of the generator.

\subsection{Answer Generation and Post-Hoc Correction}
We evaluate three different generative pipelines against the LongEval Organizers' naive baseline.

\subsubsection{DS@GT Hybrid Baseline}
This architecture represents a standard RAG implementation. The top-$k$ chunks retrieved by the BM25+BGE hybrid ranker are injected directly into the prompt for the Gemma-4-31B instruction-tuned model. The model synthesizes an initial answer and outputs an array of cited document indices.

\subsubsection{Corrective RAG and CiteFix}
\begin{figure}[h]
    \centering
    \includegraphics[width=1.1\textwidth]{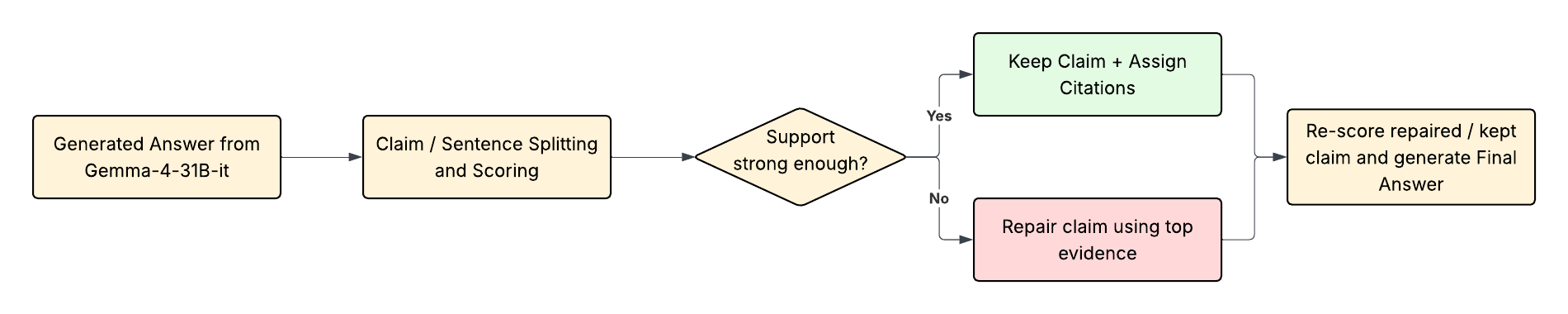}
    \caption{CiteFix Evidence Validation Repair Pipeline}
    \label{fig:CiteFix-Method}
\end{figure}

Our primary research pipeline introduces active interventions to the Hybrid Baseline.
\begin{enumerate}
    \item Pre-Generation (CRAG): Before generation, a lightweight cross-encoder (cross-encoder/ms-marco-MiniLM-L-6-v2) evaluates retrieved chunks. Chunks below the entailment threshold are dropped from the supplied context.
    \item Post-Generation (CiteFix): After generation, we use CiteFix to parse generated claims against cited document chunks. Citations lacking entailment to support the generated claims are pruned. Figure \ref{fig:CiteFix-Method} shows the steps of the CiteFix pipeline.
\end{enumerate}

\subsubsection{Frontier Model Benchmarks}
To establish an upper bound on long-context parametric reasoning, we bypass scoring and chunking entirely. We prompt ChatGPT-5.5 Thinking and Claude Opus 4.7 Thinking to synthesize answers directly from the raw text of the 10 candidate documents.

\subsection{Reference-Free RAGAs Evaluation}

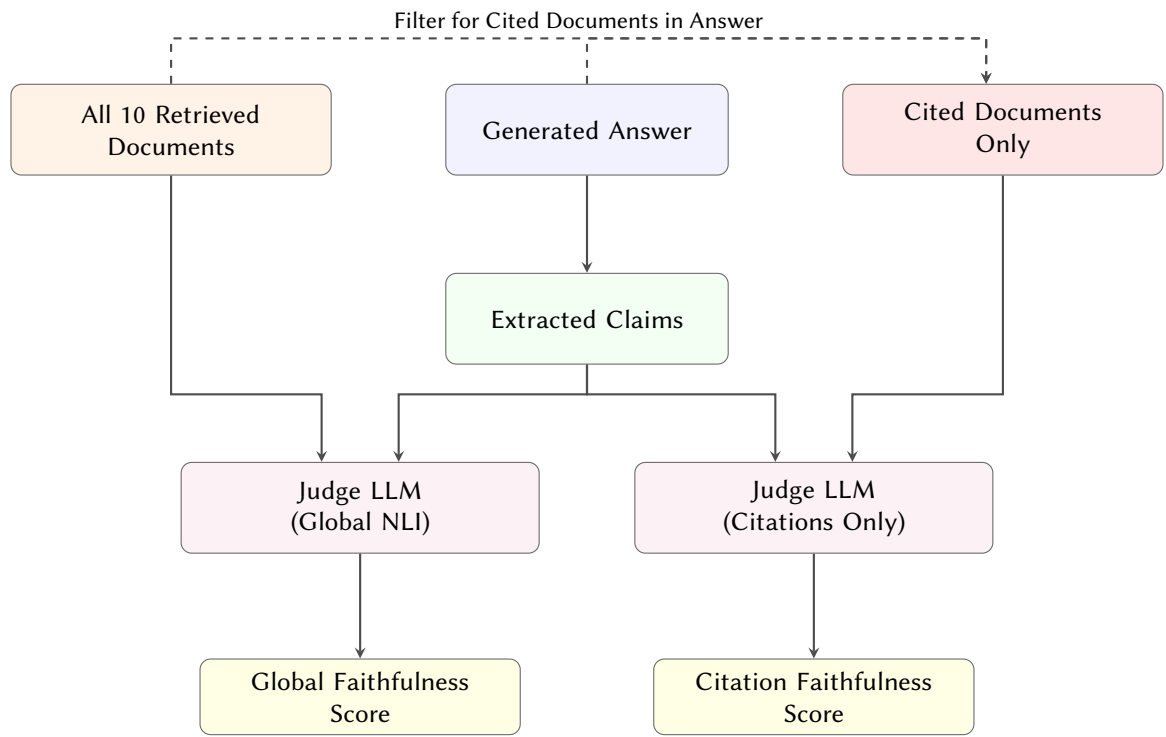
\begin{figure*}[htbp]
\centering
\begin{tikzpicture}[
    box/.style={rectangle, draw=black!60, fill=blue!5, text width=3.5cm, text centered, rounded corners, minimum height=1.2cm, font=\small},
    judge/.style={rectangle, draw=black!60, fill=green!5, text width=3.5cm, text centered, rounded corners, minimum height=1.2cm, font=\small},
    docall/.style={rectangle, draw=black!60, fill=orange!10, text width=4cm, text centered, rounded corners, minimum height=1.2cm, font=\small},
    doccite/.style={rectangle, draw=black!60, fill=red!10, text width=4cm, text centered, rounded corners, minimum height=1.2cm, font=\small},
    calc/.style={rectangle, draw=black!60, fill=purple!5, text width=4.5cm, text centered, rounded corners, minimum height=1.2cm, font=\small},
    score/.style={rectangle, draw=black!60, fill=yellow!10, text width=4cm, text centered, rounded corners, minimum height=1cm, font=\small},
    arrow/.style={thick,->,>=stealth, draw=black!70}]

    \node (ans) [box] at (0, 0) {Generated Answer};
    \node (alldocs) [docall] at (-5.5, 0) {All 10 Retrieved\\Documents};
    \node (citedocs) [doccite] at (5.5, 0) {Cited Documents\\Only};

    \node (claims) [judge] at (0, -2.5) {Extracted Claims};

    \node (global_nli) [calc] at (-3, -5) {Judge LLM\\(Global NLI)};
    \node (cite_nli) [calc] at (3, -5) {Judge LLM\\(Citations Only)};

    \node (global_score) [score] at (-3, -7.5) {Global Faithfulness\\Score};
    \node (cite_score) [score] at (3, -7.5) {Citation Faithfulness\\Score};

    \draw [arrow] (ans) -- (claims);
    \draw [arrow] (alldocs.south) |- (-5.5, -3.5) -| (global_nli.130);
    \draw [arrow] (citedocs.south) |- (5.5, -3.5) -| (cite_nli.50);
    \draw [arrow] (claims.south) |- (0, -3.5) -| (global_nli.50);
    \draw [arrow] (claims.south) |- (0, -3.5) -| (cite_nli.130);

    \draw [arrow] (global_nli) -- (global_score);
    \draw [arrow] (cite_nli) -- (cite_score);

    \draw [arrow, dashed] (alldocs.north) -- ++(0,0.6) -| node[pos=0.25, above, font=\footnotesize] {Filter for Cited Documents in Answer} (citedocs.110);
    \draw [arrow, dashed] (ans.north) -- ++(0,0.6) -| (citedocs.110);

\end{tikzpicture}
\caption{RAGAs Faithfulness workflows. Global Faithfulness verifies claims extracted from the generated answer against the entire 10-document context array to penalize ungrounded statements. Citation Faithfulness filters for only cited documents, forcing the judge to verify claims strictly against the documents cited in the answer.}
\label{fig:faithfulness-workflow}
\end{figure*}

\begin{figure*}[htbp]
\centering
\begin{tikzpicture}[
    box/.style={rectangle, draw=black!60, fill=blue!5, text width=3.5cm, text centered, rounded corners, minimum height=1.2cm, font=\small},
    judge/.style={rectangle, draw=black!60, fill=green!5, text width=4.5cm, text centered, rounded corners, minimum height=1.2cm, font=\small},
    query/.style={rectangle, draw=black!60, fill=orange!5, text width=3cm, text centered, rounded corners, minimum height=1.2cm, font=\footnotesize},
    calc/.style={rectangle, draw=black!60, fill=purple!5, text width=4.5cm, text centered, rounded corners, minimum height=1.2cm, font=\small},
    score/.style={rectangle, draw=black!60, fill=yellow!10, text width=4cm, text centered, rounded corners, minimum height=1.2cm, font=\small},
    arrow/.style={thick,->,>=stealth, draw=black!70}]

    \node (ans) [box] at (-3, 0) {Generated Answer};
    \node (orig) [box, fill=red!10] at (3, 0) {Original Query};
    
    \node (llm) [judge] at (-3, -2.5) {Judge LLM\\(Generate $N$ Queries)};
    
    \node (q1) [query] at (-5, -5) {Hypothetical\\Query 1};
    \node (qn) [query] at (-1, -5) {Hypothetical\\Query $N$};
    
    \node (cos) [calc] at (0, -7.5) {Cosine Similarity};
    
    \node (score) [score] at (0, -9.5) {Answer Relevancy\\Score};

    \draw [arrow] (ans) -- (llm);
    \draw [arrow] (llm) -- (q1);
    \draw [arrow] (llm) -- (qn);
    \draw [thick, dotted, draw=black!50] (q1) -- (qn);
    
    \draw [arrow] (q1.south) |- (-5, -6.2) -| (cos.130);
    \draw [arrow] (qn.south) |- (-1, -6.2) -| (cos.130);
    \draw [arrow] (orig.south) |- (3, -6.2) -| (cos.50);
    
    \draw [arrow] (cos) -- (score);

\end{tikzpicture}
\caption{RAGAs Answer Relevancy workflow. Only the generated answer is provided to the judge model in order to generate counterfactual queries. These are compared to the original user query via embedding cosine similarity.}
\label{fig:answer-relevancy}
\end{figure*}
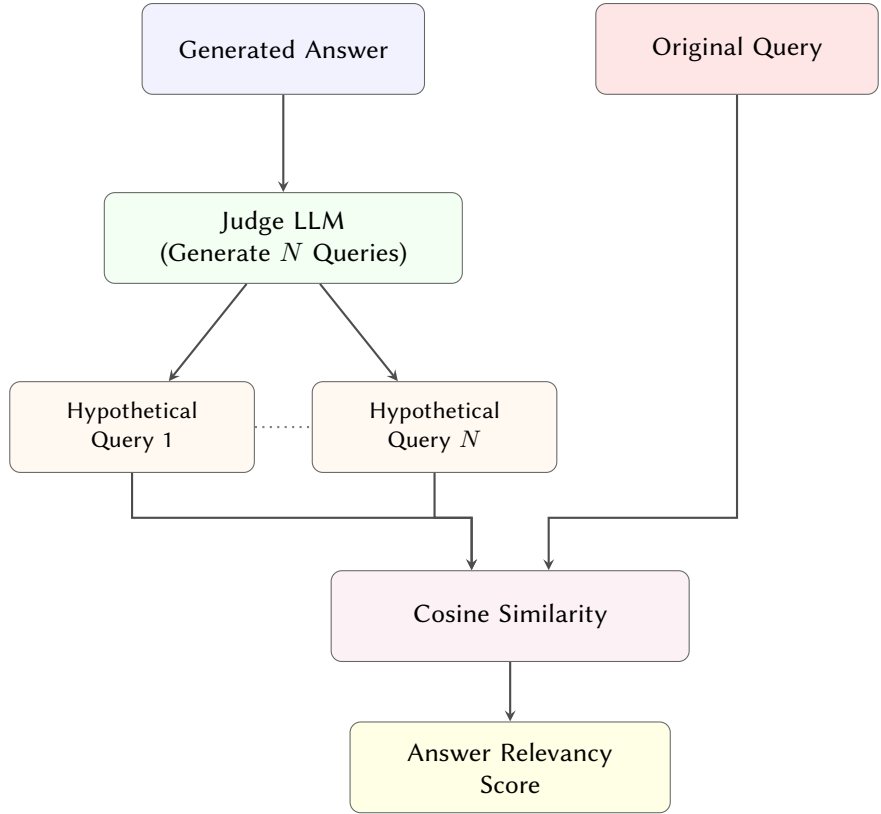

In the absence of a gold QA set with ground truth answers in the development phase, we utilized an LLM-as-judge setup (Claude Sonnet 4.6) via the RAGAs framework \cite{es-etal-2024-ragas}. The evaluation calculated the faithfulness and relevancy metrics using multi-step prompting algorithms with the judge LLM. The scoring workflows are presented in Figures \ref{fig:faithfulness-workflow} and \ref{fig:answer-relevancy}. 

\paragraph{Global Faithfulness} This metric measures the factual consistency of the generated response against the 10 pre-retrieved documents. Using RAGAs, the generated answer is decomposed into discrete, standalone claims. The judge LLM (Claude Sonnet 4.6) then decides if claims in the generated answer are grounded within the entirety of the supplied documents. The final metric is the ratio of grounded statements to total statements in answers on the evaluation set.

\paragraph{Citation Faithfulness} While Global Faithfulness evaluates the entire task context, Citation Faithfulness isolates the documents claimed to have been used at generation. During this evaluation pass, answer claims are evaluated against only cited documents. Large gaps between Global Faithfulness and Citation Faithfulness reveal a failure mode where grounded answers are produced but falsify or otherwise hallucinate the citation.

\paragraph{Answer Relevancy} This metric quantifies the completeness of the answer with respect to the query. Rather than use semantic similarity of question and answer, this metric uses the judge model to produce counterfactual queries based on only the generated answer. The similarity of the generated queries and the original query is then used to score each answer. This metric rewards query-aligned answers, but systematically penalizes safe abstention when the generator receives no context or context not relevant to the original query.

\subsection{Official Task Evaluation Metrics} While we performed our own suite of evaluation tests, the LongEval Organizers also evaluated submissions against a hidden gold set of answers. The headline metrics evaluate output quality using ROUGE and BERT scores. ROUGE measures structural lexical overlap by calculating the longest common subsequence (LCS) and n-gram intersection between submitted and gold answers. Conversely, BERT Score evaluates semantic equivalence by calculating cosine similarity of token embeddings. Higher metric scores indicate the generated answer shares higher lexical or semantic similarity with the gold answer.

Additionally, the organizers calculate citation precision, which evaluates whether the documents cited by our pipeline match those selected in the gold set. Let $C_q$ be the set of documents cited by our pipeline for a given user query $q$, and let $R_q$ be the set of documents labeled relevant in the gold standard. Citation precision calculates the ratio of correctly cited relevant documents to the total number of cited documents ($|C_q \cap R_q| / |C_q|$). The final score is averaged over all queries. The organizers provide detailed metrics covering natural language coherency, consistency, and clarity, as well as a variety of other retrieval heuristics measuring the prompt in comparison to the input query. The full list of the results including the additional IR axiom metrics can be viewed in the Appendix.

\section{Results}

Here, we report the official LongEval-RAG evaluation metrics and our own reference-free RAGAs diagnostics for each of our pipelines.

\subsection{Official Task Metrics}
Tables \ref{tab:main-results} and \ref{tab:additional-metric-breakdown} detail aggregate performance of the pipelines on the official task metrics. The frontier model runs establish a high upper bound for natural language generation. GPT-5.5 achieved the highest similarity scores with a $\text{ROUGE}_{F1}$ of 0.188 and a $\text{BERT}_{F1}$ of 0.227. Claude Opus 4.7 achieved a perfect Citation Precision (1.000) and the highest Nugget Coverage (0.443), the best performance in selecting the correct document IDs from each pre-retrieved set of documents.

We see that our hybrid chunk reranking benchmark significantly outperformed the Organizer's Naive Baseline across all lexical and semantic metrics. Our corrective intervention pipeline using CRAG+CiteFix experienced a drop in both ROUGE and BERT scores compared to the hybrid baseline. This is an artifact of the abstention behavior which suppresses generation of ungrounded text.

\begin{table*}[h]
\centering
\small
\setlength{\tabcolsep}{6pt}
\rowcolors{2}{gray!10}{white}
\begin{tabular}{lccccc}
\toprule
\textbf{System} & $\mathbf{ROUGE_{F1}}$ & $\mathbf{BERT_{F1}}$ & \textbf{Precision} & \textbf{Nugget Cov.} & \textbf{TFC1} \\
\midrule
(LongEval Lab) Naive Baseline & 0.082 & -0.118 & 0.200 & 0.124 & -0.758 \\
(DS@GT) Hybrid Baseline & 0.132 & 0.118 & 0.766 & 0.210 & -0.322 \\
CRAG+CiteFix & 0.126 & 0.076 & 0.692 & 0.152 & -0.505 \\
GPT-5.5 Thinking & \textbf{0.188} & \textbf{0.227} & 0.904 & 0.308 & \textbf{0.434} \\
Claude Opus 4.7 Thinking & 0.169 & 0.157 & \textbf{1.000} & \textbf{0.443} & 0.017 \\
\bottomrule
\end{tabular}
\caption{LongEval Task 4 results across automatic and manually curated runs. Manual systems provide an approximate upper bound for natural language metrics.}
\label{tab:main-results}
\end{table*}

Table \ref{tab:clarity-metrics} shows system performance on secondary metrics evaluating sentence-level consistency and clarity based on TREC-AutoJudge Axioms \cite{merker2025axioms, farzi2026autojudge}. Claude Opus 4.7 produced the most coherent output ($\text{COH1} = 0.857$). The Naive Baseline scored well on CLAR2, indicating high readability despite poor accuracy.

\begin{table*}[h]
\centering
\small
\setlength{\tabcolsep}{6pt}
\rowcolors{2}{gray!10}{white}
\begin{tabular}{lccccc}
\toprule
\textbf{System} & $\mathbf{ROUGE_1}$ & $\mathbf{ROUGE_2}$ & $\mathbf{ROUGE_L}$ & $\mathbf{BERT_P}$ & $\mathbf{BERT_R}$ \\
\midrule
(LongEval Lab) Naive Baseline & 0.142 & 0.023 & 0.082 & -0.313 & 0.090 \\
(DS@GT) Hybrid Baseline  & 0.207 & 0.023 & 0.132 & 0.117 & 0.117 \\
CRAG+CiteFix & 0.201 & 0.024 & 0.126 & 0.064 & 0.086 \\
GPT-5.5 Thinking & \textbf{0.292} & \textbf{0.062} & \textbf{0.188} & \textbf{0.242} & 0.210 \\
Claude Opus 4.7 Thinking & 0.284 & 0.057 & 0.169 & 0.097 & \textbf{0.217} \\
\bottomrule
\end{tabular}
\caption{Detailed gold-answer similarity metrics. R-1 and R-2 are Rouge Scores for lexical similarity between generated and gold set on word-one-grams and bi-grams respectively. R-L is Rouge Score for longest overlapping phrase. BERT-P and BERT-R, are BERT precision and recall respectively for all queries. GPT-5.5 manual outputs achieve the strongest $ROUGE_1$, $ROUGE_2$, $ROUGE_L$, and $BERT_P$ scores, while Claude Opus 4.7 Thinking manual outputs achieve the highest $BERT_R$ score.}
\label{tab:additional-metric-breakdown}
\end{table*}

\begin{table*}[h]
\centering
\small
\setlength{\tabcolsep}{6pt}
\rowcolors{2}{gray!10}{white}
\begin{tabular}{lrrrr}
\toprule
\textbf{System} & \textbf{COH1} & \textbf{COH2} & \textbf{CLAR1} & \textbf{CLAR2} \\
\midrule
(LongEval Lab) Naive Baseline & -0.150 & -0.064 & -0.093 & 0.678 \\
(DS@GT) Hybrid Baseline  & 0.024 & 0.020 & -0.006 & -0.030 \\
CRAG+CiteFix & -0.021 & \textbf{0.021} & 0.013 & -0.183 \\
GPT-5.5 Thinking & 0.427 & 0.004 & \textbf{0.014} & 0.239 \\
Claude Opus 4.7 Thinking & \textbf{0.857} & -0.119 & 0.013 & \textbf{0.818} \\
\bottomrule
\end{tabular}
\caption{Claude Opus 4.7 Thinking manual responses are strongly preferred on sentence-level coherency and clarity (COH1 and CLAR2). GPT-5.5 Thinking and CRAG+CiteFix approach lead on secondary metrics (CLAR1 and COH2 respectively). The organizer naive baseline is weak on coherence but scores highly on CLAR2.}
\label{tab:clarity-metrics}
\end{table*}

\begin{table*}[h]
\centering
\small
\setlength{\tabcolsep}{6pt}
\rowcolors{2}{gray!10}{white}
\begin{tabular}{lccc}
\toprule
\textbf{System} & \textbf{Global Faithfulness} & \textbf{Citation Faithfulness} & \textbf{Answer Relevancy} \\
\midrule
Hybrid Baseline & 0.741 & 0.741 & 0.531 \\
CRAG+CiteFix & \textbf{0.784} & \textbf{0.758} & 0.476 \\
GPT-5.5 Thinking & 0.559 & 0.417 & \textbf{0.756} \\
Claude Opus 4.7 Thinking & 0.553 & 0.528 & 0.646 \\ 
\bottomrule
\end{tabular}
\caption{RAGAs automated evaluation metrics across generation architectures. CRAG+CiteFix maximizes factual grounding and attribution accuracy, while frontier models (Opus, GPT-5.5) prioritize answer relevancy at the expense of strict citation integrity.}
\label{tab:RAGAs-metrics}
\end{table*}

\subsection{RAGAs Evaluation Metrics}
Table \ref{tab:RAGAs-metrics} shows the results of our reference-free RAGAs evaluation. These metrics evaluate strictly for attribution integrity and grounding of claims rather than lexical overlaps or semantic similarity. Our CRAG+CiteFix pipeline maximized faithfulness to the supplied context, achieving a Global Faithfulness of 0.784 and a Citation Faithfulness of 0.758. Conversely, the frontier models performed quite poorly on these evaluations. Faithfulness scores for both GPT-5.5 Thinking and Claude Opus 4.7 Thinking collapsed despite producing answers that were highly relevant to the queries. This indicates a failure mode where the models output highly relevant answers without extracting their claims from the cited documents.

\section{Discussion}

We observe a tradeoff between several performance metrics and our RAGAs-based faithfulness and citation accuracy evaluation when comparing structured corrective pipelines with unconstrained frontier models. While the frontier models (GPT-5.5, Opus 4.7) achieved higher official leaderboard scores, our RAGAs diagnostics (Table \ref{tab:RAGAs-metrics}) indicate a failure to ground the answers in the retrieved context. Faced with noisy context, the frontier models generated answers predominantly from pre-trained knowledge.

For example, GPT-5.5 generally led the leaderboard scores and had the highest Answer Relevancy of 0.756, but exhibited extremely low Citation Faithfulness at 0.417 indicating low reliance on cited contexts in answers. From the leaderboard precision of 0.904 and manual trace audits, we find that while GPT-5.5 is highly capable of identifying the relevant document(s) within a noisy context window, the details of generated answers typically did not rely on that same context.

The corrective pipeline achieved Global Faithfulness of 0.784 and Citation Faithfulness of 0.758, which were leading on our submissions. However, this triggered penalties on some of the leaderboard metrics. When retrieved chunks lacked the answer, our corrective pipeline would abstain to answer rather than generate an answer from parametric memory. These abstentions are penalized on metrics reliant on relevancy or n-gram overlaps despite exhibiting a more faithful failure mode than parametric memory leaks. This highlights a challenge in trustworthy RAG benchmarking where chosen evaluation metrics may reward failure modes which hallucinate citations to maintain conversational fluency over safer abstentions.

To evaluate the statistical significance of the CRAG + CiteFix pipeline, we conducted a Wilcoxon signed-rank test on the paired RAGAs evaluation scores across the 47 test queries. The results indicated that the corrective interventions did not yield a statistically significant improvement to citation faithfulness or answer relevancy.

\section{Future Work}

A few limitations to our evaluation should be noted. While our frontier model submissions score well on lexical similarity metrics, they score poorly on faithfully grounding answers in the retrieved evidence. Conversely, our Gemma-4-31B pipelines exhibit relatively high faithfulness but much lower relevance and citation precision. As our prompting strategy did not require answers to be strictly extractive, instead requesting synthesis across relevant passages, we posit that forcing output text to be extractive would greatly improve lexical similarity and potentially induce more faithful answering from the higher capacity models, if at a penalty to clarity and coherence.

An immediate next step for this work is to verify that the CRAG and CiteFix corrective interventions successfully improve answer grounding across families and sizes of generators. Our initial results do not indicate a statistically significant increase in faithfulness over the baseline. Further exploration of NLI model use for scientific claim verification is needed to understand if the null result is driven by NLI brittleness rather than applicability of post-hoc pruning in this domain. Additionally, the lack of a CRAG+CiteFix submission to the leaderboard using GPT-5.5 Thinking or Claude Opus 4.7 Thinking was due to practical submission limit and time constraints within the evaluation period. We encourage release of the hidden gold set query answers to facilitate expanded ablation studies.

Our RAGAs evaluations are not without limitations. A single, uncalibrated reference-free LLM judge was used to evaluate answer faithfulness and relevancy. Future evaluations will deploy multi-judge ensembles, enabling the evaluation of intraclass correlations across judges and mitigation of known evaluator biases such as self-preference or verbosity.

Finally, the described RAG systems and RAGAs evaluators should be extended across multiple snapshots within the LongEval corpus. The current evaluation focused on a static payload. Applying these pipelines across a longitudinal dataset would assess QA performance under the types of temporal drift which penalize use of point-in-time parametric memory. Additionally, the threshold hyperparameters for CRAG and CiteFix are tuned for a specific generation task. The performance of these techniques over multiple snapshots may degrade, similar to standard retrievers and generators. Longitudinal evaluation of these systems should explore adaptive tuning of the active interventions, as well as assess how corpus shifts impact the faithfulness and relevancy judgments of the underlying RAGAs LLM judge. As our faithfulness and relevancy metrics are tied to an LLM-as-a-judge method, temporal bias should be evaluated and compared to other known judge biases when interpreting longitudinal RAGAs scores.

\section{Conclusions}

We find a divergence between traditional language metrics and factual grounding in RAG. While unconstrained frontier models generally achieved better ROUGE and BERT score with query-aligned answers, our diagnostics show this comes from utilization of parametric memory rather than retrieved context. Conversely, we find that deployment of active interventions pre- and post-generation do improve structural honesty in RAG QA systems. Ultimately, securing RAG QA pipelines in scientific domains rely on both strict grounding to evidence and overall quality of answer. We therefore recommend adoption of evaluation metrics that reward structural safety and proper failure modes within RAG systems. \newpage

\section*{Acknowledgments}

We thank the Data Science at Georgia Tech Applied Research Competitions (DS@GT ARC) group for their support. This research was supported in part through research cyberinfrastructure resources and services provided by the Partnership for an Advanced Computing Environment (PACE) at the Georgia Institute of Technology, Atlanta, Georgia, USA \cite{PACE}. 

\section*{Declaration on Generative AI}
During the preparation of this work, the authors used Gemini 3.1 Pro in order to aid in formatting, grammar, and spell checking. After using these tools, the authors reviewed and edited the content as needed and take full responsibility for the publication’s content.

\bibliography{main} \newpage

@article{yan2024crag,
  title = {Corrective Retrieval Augmented Generation},
  author = {Yan, Shi-Qi and Gu, Jia-Chen and Zhu, Yun and Ling, Zhen-Hua},
  journal = {arXiv preprint arXiv:2401.15884},
  year = {2024},
  url = {https://arxiv.org/abs/2401.15884}
}

@inproceedings{maheshwari2025citefix,
  title = {CiteFix: Enhancing {RAG} Accuracy Through Post-Processing Citation Correction},
  author = {Maheshwari, Harsh and others},
  booktitle = {Proceedings of the 63rd Annual Meeting of the Association for Computational Linguistics: Industry Track},
  year = {2025},
  url = {https://aclanthology.org/2025.acl-industry.23/}
}

@inproceedings{lewis2020rag,
  title = {Retrieval-Augmented Generation for Knowledge-Intensive NLP Tasks},
  author = {Lewis, Patrick and Perez, Ethan and Piktus, Aleksandra and Petroni, Fabio and Karpukhin, Vladimir and Goyal, Naman and K{\"u}ttler, Heinrich and Lewis, Mike and Yih, Wen-tau and Rockt{\"a}schel, Tim and Riedel, Sebastian and Kiela, Douwe},
  booktitle = {Advances in Neural Information Processing Systems},
  year = {2020}
}

@inproceedings{keller2024eval,
author = {Keller, J\"{u}ri and Breuer, Timo and Schaer, Philipp},
title = {Evaluation of Temporal Change in IR Test Collections},
year = {2024},
isbn = {9798400706813},
publisher = {Association for Computing Machinery},
address = {New York, NY, USA},
url = {https://doi.org/10.1145/3664190.3672530},
doi = {10.1145/3664190.3672530},
booktitle = {Proceedings of the 2024 ACM SIGIR International Conference on Theory of Information Retrieval},
pages = {3–13},
numpages = {11},
location = {Washington DC, USA},
series = {ICTIR '24}
}

@inproceedings{liu2023evaluating,
    title = "Evaluating Verifiability in Generative Search Engines",
    author = "Liu, Nelson  and
      Zhang, Tianyi  and
      Liang, Percy",
    editor = "Bouamor, Houda  and
      Pino, Juan  and
      Bali, Kalika",
    booktitle = "Findings of the Association for Computational Linguistics: EMNLP 2023",
    month = dec,
    year = "2023",
    address = "Singapore",
    publisher = "Association for Computational Linguistics",
    url = "https://aclanthology.org/2023.findings-emnlp.467/",
    doi = "10.18653/v1/2023.findings-emnlp.467",
    pages = "7001--7025"
}

@inproceedings{gao-etal-2023-enabling,
    title = "Enabling Large Language Models to Generate Text with Citations",
    author = "Gao, Tianyu  and
      Yen, Howard  and
      Yu, Jiatong  and
      Chen, Danqi",
    editor = "Bouamor, Houda  and
      Pino, Juan  and
      Bali, Kalika",
    booktitle = "Proceedings of the 2023 Conference on Empirical Methods in Natural Language Processing",
    month = dec,
    year = "2023",
    address = "Singapore",
    publisher = "Association for Computational Linguistics",
    url = "https://aclanthology.org/2023.emnlp-main.398/",
    doi = "10.18653/v1/2023.emnlp-main.398",
    pages = "6465--6488"
}

@manual{PACE,
  title  = {{P}artnership for an {A}dvanced {C}omputing {E}nvironment ({PACE})},
  author = {{PACE}},
  url    = {http://www.pace.gatech.edu},
  year   = {2017}
}

@inproceedings{merker2025axioms,
  author    = {J. H. Merker and M. Fr\"{o}be and B. Stein and M. Potthast and M. Hagen},
  title     = {Axioms for retrieval-augmented generation},
  booktitle = {Proceedings of the 2025 International ACM SIGIR Conference on Innovative Concepts and Theories in Information Retrieval, ICTIR 2025},
  pages     = {67--77},
  year      = {2025},
  publisher = {ACM},
  address   = {Padua, Italy},
  doi       = {10.1145/3731120.3744601},
  url       = {https://doi.org/10.1145/3731120.3744601}
}

@misc{farzi2026autojudge,
  author    = {N. Farzi and T. Hagen and E. Yang and M. Fr\"{o}be and R. Pradeep and H. A. Rahmani and X. Wang and O. Zendel and M. P. L. Dietz},
  title     = {Auto-judge: A cross-task benchmark for comparing llm judges for citation-grounded rag systems},
  year      = {2026}
}

@inproceedings{es-etal-2024-ragas,
    title = "{RAGA}s: Automated Evaluation of Retrieval Augmented Generation",
    author = "Es, Shahul  and
      James, Jithin  and
      Espinosa Anke, Luis  and
      Schockaert, Steven",
    editor = "Aletras, Nikolaos  and
      De Clercq, Orphee",
    booktitle = "Proceedings of the 18th Conference of the European Chapter of the Association for Computational Linguistics: System Demonstrations",
    month = mar,
    year = "2024",
    address = "St. Julians, Malta",
    publisher = "Association for Computational Linguistics",
    url = "https://aclanthology.org/2024.eacl-demo.16/",
    doi = "10.18653/v1/2024.eacl-demo.16",
    pages = "150--158"
}

@article{Knoth2023-zi,
  title  = {CORE: A global aggregation service for open access papers},
  author = {Knoth, Petr and Herrmannova, Drahomira and Cancellieri, Matteo
  and Anastasiou, Lucas and Pontika, Nancy and Pearce, Samuel and
  Gyawali, Bikash and Pride, David},
  journal = {Nature Scientific Data},
  volume = 10,
  number = 1,
  pages  = 366,
  month  = jun,
  year   = 2023,
  language = {en}
}

@inproceedings{lncs_overview_longeval_2026,
 author       = {Sarah Bouaraba and Timo Breuer and
                 Matteo Cancellieri and
                 Alaa El{-}Ebshihy and
                 Maik Fr{\"{o}}be and
                 Petra Galusc{\'{a}}kov{\'{a}} and
                 Lorraine Goeuriot and
                 Gabriel Iturra{-}Bocaz and
                 J{\"{u}}ri Keller and
                 Petr Knoth and
                 Andreas Konstantin Kruff and
                 Philippe Mulhem and
                 Florina Piroi and
                 David Pride and
                 Philipp Schaer and
                 Didier Schwab},
 title     = {Overview of the CLEF 2026 LongEval Lab on Longitudinal Evaluation of Model Performance},
 booktitle = {Experimental {IR} Meets Multilinguality, Multimodality, and Interaction. Proceedings of the Seventeenth International Conference of the {CLEF} Association ({CLEF} 2026)},
 editor    = {Hagen, Matthias and Potthast, Martin and Stein, Benno and Schaer, Philipp and Zangerle, Eva and MacAvaney, Sean and Stru{\ss}, Julia Maria and S\'{a}nchez Salido, Eva and Barr\'{o}n-Cede\~{n}o, Alberto and Garc\'{i}a Seco de Herrera, Alba},
 publisher = {Springer},
 series    = {Lecture Notes in Computer Science (LNCS)},
 year      = {2026}
}

@inproceedings{ceur_overview_longeval_2026,
 author       = {Sarah Bouaraba and Timo Breuer and
                 Matteo Cancellieri and
                 Alaa El{-}Ebshihy and
                 Maik Fr{\"{o}}be and
                 Petra Galusc{\'{a}}kov{\'{a}} and
                 Lorraine Goeuriot and
                 Gabriel Iturra{-}Bocaz and
                 J{\"{u}}ri Keller and
                 Petr Knoth and
                 Andreas Konstantin Kruff and
                 Philippe Mulhem and
                 Florina Piroi and
                 David Pride and
                 Philipp Schaer and
                 Didier Schwab},
 title     = {Extended Overview of the CLEF 2026 LongEval Lab on Longitudinal Evaluation of Model Performance},
 booktitle = {{CLEF} 2026 Working Notes},
 editor    = {S\'{a}nchez Salido, Eva and Barr\'{o}n-Cede\~{n}o, Alberto and Garc\'{i}a Seco de Herrera, Alba and MacAvaney, Sean and Stru{\ss}, Julia Maria},
 publisher = {CEUR-WS.org},
 series    = {CEUR Workshop Proceedings},
 year      = {2026}
}

\section{Appendix}

\subsection{Answer Generation Prompt}

For each query, the LLM receives both the user scientific query and the CRAG-filtered evidence block for the research based approaches (excluded for baseline). The prompt instructs the model to produce a concise, paraphrased answer that synthesizes evidence from the retrieved document chunks without directly copying. The exact prompt template used is below:

\begin{quote}
\begin{verbatim}
Question:
{query}

CRAG-filtered evidence:
{evidence_block}

Write a direct answer in 1-3 sentences, about 30-65 words total.

Rules:
- No bullets, markdown, or inline citations.
- Do not mention passage numbers or evidence labels.
- Synthesize the answer rather than copying source text.
- Even if evidence is partial, give the strongest concise answer
  supported by the evidence.
- Write in English only.
- Use standard English spellings for chemical names and technical terms.
\end{verbatim}
\end{quote}

\subsection{TREC-AutoJudge IR Axiom Results}
\begin{table*}[h]
\centering
\scriptsize
\setlength{\tabcolsep}{4pt}
\resizebox{\textwidth}{!}{
\rowcolors{2}{gray!10}{white}
\begin{tabular}{lrrrrr}
\toprule
\textbf{Metric} & \textbf{Naive Baseline} & \textbf{Hybrid Baseline} & \textbf{CRAG+CiteFix} & \textbf{GPT-5.5 Thinking} & \textbf{Claude Opus 4.7 Thinking} \\
\midrule
GEN-TFC1    & -0.758 & -0.322 & -0.505 & \textbf{0.434} & 0.017 \\
GEN-LNC1    &  0.000 &  0.001 & \textbf{0.005} & 0.000 & 0.000 \\
GEN-REG     & -0.184 & -0.126 & -0.257 & \textbf{0.345} & 0.065 \\
GEN-AND     & \textbf{0.737} & -0.263 & -0.242 & -0.157 & -0.136 \\
GEN-DIV     & \textbf{0.626} &  0.223 &  0.406 & -0.512 & -0.290 \\
GEN-STMC1   & -0.662 & -0.112 & -0.364 & \textbf{0.501} & -0.124 \\
GEN-STMC2   &  0.041 &  0.018 & -0.008 & 0.018 & \textbf{0.042} \\
GEN-PROX1   & \textbf{0.137} &  0.006 &  0.002 & 0.014 & -0.007 \\
GEN-PROX2   & -0.083 & -0.008 & -0.004 & -0.029 & \textbf{0.055} \\
GEN-PROX3   & \textbf{0.116} &  0.000 &  0.001 & 0.001 & -0.015 \\
GEN-PROX4   & \textbf{0.136} &  0.000 &  0.000 & 0.002 & 0.013 \\
GEN-PROX5   & \textbf{0.050} &  0.000 &  0.006 & 0.025 & 0.019 \\
GEN-ASL     &  0.000 &  0.000 &  0.000 & 0.000 & 0.000 \\
GEN-TF-LNC  & -0.005 & -0.010 & -0.013 & \textbf{0.009} & -0.008 \\
\bottomrule
\end{tabular}}
\caption{Results for TREC-AutoJudge IR axiom metrics. The GEN-* metrics evaluate proximity properties of generated answers with respect to the input query. Higher values indicate better results.}
\label{tab:appendix-gen-axioms}
\end{table*}

\begin{table*}[t!]
\centering
\scriptsize
\setlength{\tabcolsep}{4pt}
\resizebox{\textwidth}{!}{
\rowcolors{2}{gray!10}{white}
\begin{tabular}{lrrrrr}
\toprule
\textbf{Metric} & \textbf{Naive Baseline} & \textbf{Hybrid Baseline} & \textbf{CRAG+CiteFix} & \textbf{GPT-5.5 Thinking} & \textbf{Claude Opus 4.7 Thinking} \\
\midrule
COH1-0.75       & -0.150 &  0.024 & -0.021 & 0.427 & \textbf{0.857} \\
COH2-0.75       & -0.064 &  0.020 & \textbf{0.021} & 0.004 & -0.119 \\
COV1-SE-0.5     & \textbf{0.859} & -0.150 & -0.047 & -0.404 & 0.165 \\
COV2-SE-0.5     & \textbf{0.363} & -0.024 &  0.005 & -0.051 & -0.107 \\
COV3-SE-0.5     & \textbf{0.018} & -0.017 & -0.003 & -0.025 & -0.032 \\
CONS3-0.5       & \textbf{0.005} &  0.003 & \textbf{0.005} & 0.003 & \textbf{0.005} \\
CORR1-0.75      & \textbf{0.001} & \textbf{0.001} & \textbf{0.001} & \textbf{0.001} & \textbf{0.001} \\
CLAR1-0.5       & -0.093 & -0.006 & \textbf{0.013} & \textbf{0.013} & \textbf{0.013} \\
CLAR2-0.5       &  0.678 & -0.030 & -0.183 & 0.239 & \textbf{0.818} \\
\bottomrule
\end{tabular}}
\caption{Appendix results for TREC-AutoJudge coherence, coverage, consistency, correction, and clarity metrics.}
\label{tab:appendix-coh-cov-cons-clar}
\end{table*}

Tables \ref{tab:appendix-gen-axioms} and \ref{tab:appendix-coh-cov-cons-clar} detail our submission results on TREC-AutoJudge IR Axioms. Notably, the organizers' naive baseline performed best across several proximity and intersection-based metrics. We hypothesize that the performance gap is an artifact of the divergence between extractive and abstractive generation methods. Classical IR axioms like GEN-AND and GEN-PROX4 reward lexical overlap and term proximity. An extractive baseline will inherently satisfy these conditions. Our RAG pipelines were prompted to perform abstractive synthesis across the supplied context. This process can break lexical overlaps and alter term distances, resulting in penalty by these axioms. We acknowledge that stricter extractive prompting could improve scores on these axioms.

\end{document}